\documentclass{article}
\usepackage{spconf,amsmath,graphicx}
\usepackage{multirow}

\title{Factorized Neural Transducer for Efficient Language Model Adaptation}
\name{Xie Chen, Zhong Meng, Sarangarajan Parthasarathy, Jinyu Li
}
\address{Microsoft Speech and Language Group}
\begin{document}
\ninept
\maketitle
\begin{abstract}
In recent years, end-to-end (E2E) based automatic speech recognition (ASR) systems
have achieved great success due to their simplicity and promising performance.
Neural Transducer based models are increasingly popular in streaming E2E based ASR systems and
have been reported to outperform the traditional hybrid system in some scenarios.
However, the joint optimization of acoustic model, lexicon and language model (LM)
in neural Transducer also brings about challenges
in adapting ASR using just adaptation text. 
This drawback might prevent their potential applications in practice.
In order to address this issue,
we propose a novel model, factorized neural Transducer, by factorizing the blank
and vocabulary prediction, and adopting a standalone language model for the
vocabulary prediction.
It is expected that this factorization can transfer the improvement of the
standalone language model to the Transducer for speech recognition,
which allows various language model adaptation techniques to be applied.
We demonstrate that the proposed factorized neural Transducer yields 15.4\% to 19.4\% WER improvements when out-of-domain text data is used for language model adaptation, at the cost of a minor degradation in WER on a general test set.

\end{abstract}
\begin{keywords}
factorized neural Transducer, Transformer Transducer, language model adaptation, speech recognition
\end{keywords}
\section{Introduction}
\label{sec:intro}
In recent years, end-to-end (E2E) based models \cite{chan2016listen, battenberg2017exploring,  rao2017exploring, chiu2018state, Li18CTCnoOOV, he2019streaming, Li2019RNNT, shi2020emformer}
have attracted increasing research interest in automatic speech recognition (ASR) 
systems. Compared to traditional HMM based models, where the acoustic model, 
lexicon and language model are built and optimized separately, a single neural 
network is used in E2E models to directly predict the word sequence. 
Nowadays, neural Transducer \cite{graves2012sequence, yeh2019transformer, zhang2020transformer}  and Attention-based Encoder-Decoder 
(AED) \cite{chan2016listen, Attention-bahdanau2014, 
Attention-speech-chorowski2015} are two most popular choices for E2E based ASR systems.
AED models achieved very good performance by adopting the 
attention mechanism and fusing the acoustic and linguistic information at the early stage.
However, they are not streamable models in nature. 
There are some efforts to allow AED models to work in streaming mode, 
such as monotonic chunk-wise attention \cite{DBLP:conf/iclr/ChiuR18} and 
triggered attention \cite{DBLP:conf/icassp/MoritzHR19, wang2020reducing}.
In contrast, the neural Transducer model provides a more attractive solution for 
streaming ASR and has been reported to outperform traditional hybrid systems 
\cite{Li2020Developing, sainath2020streaming, jain2019rnn} in some scenarios.
Therefore, in this work, we mainly focus on the Transducer model in light
of the streaming scenario in practice.

However, the simplicity of E2E models also brings some sacrifice.
There are no individual acoustic and language models
in a neural Transducer. Although the predictor looks similar to a 
language model in terms of model structure and an internal 
language model \cite{variani2020hat, meng2021ilme} could be extracted from
the predictor and joint network,
it does not perform as a language model because the predictor needs to coordinate with 
the acoustic encoder closely. Hence, it is not straightforward
to utilize text-only data to adapt the Transducer model 
from the source domain to the target domain. As a result,
effective and efficient language model
adaptation remains an open research problem for E2E based ASR models.

There are continuous efforts in the speech community to address this issue.
One research direction is to adopt Text-to-Speech (TTS) techniques to synthesize audio with the target-domain text \cite{Li2020Developing, sim2019personalization, zheng2021ttsasr, deng2020ttsrnnt},
and then fine-tune the Transducer model on the synthesized audio and text pairs.
However, this approach is computationally expensive. It is not
flexible and practical for scenarios requiring fast adaption.
LM fusion is another popular choice to incorporate external language 
models trained on target-domain text, such as
shallow fusion \cite{kannan2018shallowfusion} and 
density ratio based LM integration \cite{variani2020hat, meng2021ilme, mcdermott2019densityratio, meng202ilmt, meng2021mwe}.
However, the interpolation weight is task-dependent and needs to be tuned
on dev data. The performance might be sensitive to the interpolation weight.
There are some recent efforts to fine-tune the predictor \cite{pylkkonen2021fastadapt} or the internal language model \cite{Meng2021ILMA} with an additional language model loss, and then make it behave similar to a language model.
Nevertheless, the predictor in neural Transducer is not equivalent to a language model
in nature. It needs to coordinate
with the acoustic encoder, and predict the blank to prevent outputting repetitive
word \cite{ghodsi2020statelessrnn}. 

As discussed above, most previous work on LM adaptation adopted the standard neural Transducer
architecture \cite{graves2012sequence, he2019streaming}.
In this paper, we propose a modified model architecture to explicitly 
optimize the predictor towards a standard neural language model during training. 
We name it factorized neural Transducer, which factorizes the blank and vocabulary 
prediction, allowing the vocabulary predictor to work as a standalone language model.
As a result, various language model adaptation \cite{bellegarda2004lmadapt, chen2015rnnlmadapt, li2018rnnlmadapt} techniques could be simply applied to the factorized Transducer model.
The improvement
of the standalone language model is expected to yield consistent performance
gain for speech recognition, which is similar to the effect of language model in the HMM based ASR system.
We hope this work could shed some light on the re-design of model architecture, 
by disentangling the fusion of AM and LM in E2E models
for efficient language model adaptation and customization.


\section{Neural Transducer}
\label{sec:tt}

\subsection{Neural Transducer Architecture}
\begin{figure}[hbtp]
\centering
\includegraphics[width=5.5cm]{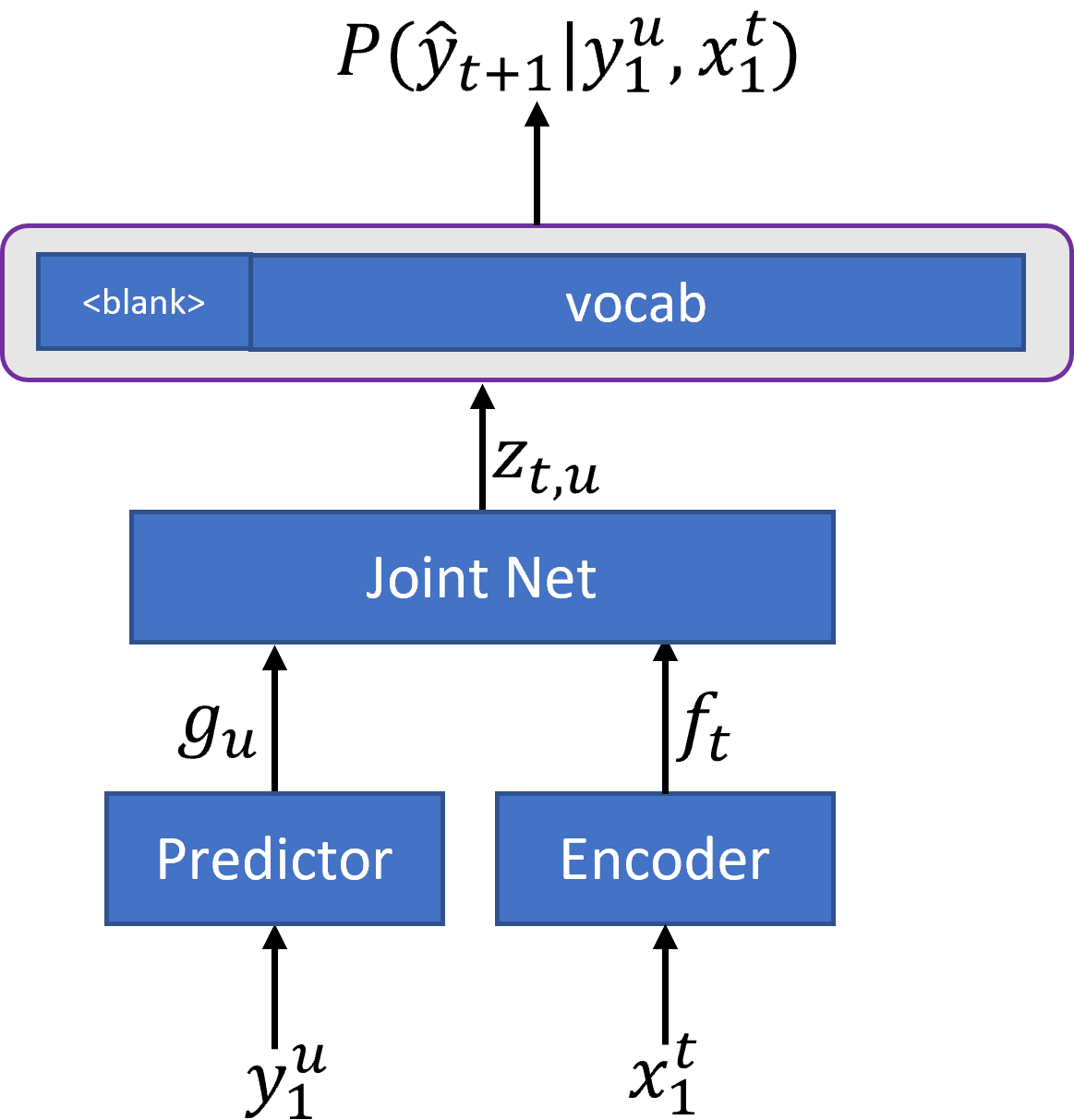}
\caption{Flowchart of neural Transducer}
\label{fig:tt}
 \vspace{-0.0cm}
\end{figure}

The neural Transducer model consists of three components, an acoustic encoder, 
a label predictor and a 
joint network, as shown in Figure \ref{fig:tt}. The encoder consumes the 
acoustic feature $\textbf{x}_1^t$ and 
generates the acoustic representation $\textbf{f}_{t}$. The predictor computes
the label representation $\textbf{g}_u$ given the history of the label sequence $\textbf{y}_1^u$.
The outputs of encoder and predictor are then combined in the joint network and
fed to the output layer to compute the probability distribution over the
output layer. The computation formulas in neural Transducer could be written as below,
\begin{eqnarray}
    \textbf{f}_{t} &=& \text{encoder}(\textbf{x}_1^{t}) \nonumber \\
    \textbf{g}_{u} &=& \text{predictor}(\textbf{y}_1^{u}) \nonumber \\
    \textbf{z}_{t, u} &=& \textbf{W}_o *\text{relu}(\textbf{f}_t+\textbf{g}_{u})    \nonumber \\
    P(\hat{y}_{t+1}|\textbf{x}_1^t, \textbf{y}_1^{u}) &=& \text{softmax}(\textbf{z}_{t, u}) 
    \label{eqn:1}
    \vspace{-0.0cm}
\end{eqnarray}

In order to address the length differences between the acoustic feature $\textbf{x}_1^T$ and label sequences $\textbf{y}_1^U$,
a special blank symbol, $\phi$, is added to the output vocabulary to 
represent a null token. Each alignment $\alpha$ contains $T+U$ output tokens,
$\hat{y}_1, ..., \hat{y}_{T+U}$, where each output token is an element of the 
set $\{\phi, \mathcal{V}\}$. The objective function of the Transducer model
is to minimize the negative log probability over all possible alignments, 
which could be written as,
\begin{equation}
    \mathcal{J}_{t} = - \log P(\textbf{y} \in \mathcal{Y}^*|\textbf{x}) = - \log \sum_{\alpha \in \beta^{-1}(\textbf{y})} P(\alpha|\textbf{x})
\label{eqn:2}
\end{equation}
where $\beta: \bar{\mathcal{Y}}^{*} \longrightarrow \mathcal{Y}^{*} $ is the function
to convert the alignment $\alpha$ to label sequence $\textbf{y}$ by removing the
blank $\phi$.

In this paper, we choose the Transformer Transducer (T-T) model as the backbone model which 
uses Transformer as the acoustic encoder, LSTM as the predictor, by considering the performance and 
computational costs. The T-T model can be trained and  evaluated efficiently as described in \cite{chen2021tt}.

\subsection{Rethinking the Blank Symbol in Neural Transducer}
\label{subsec:blank}
In the standard neural Transducer model, the predictor looks like a
language model in terms of model structure. There are some studies claiming that
an internal language model could be extracted from the Transducer model by using predictor
and joint network and excluding the blank $\phi$ connection in the output layer
\cite{variani2020hat, meng2021ilme}. Although this internal language model yields a 
relatively low perplexity \cite{meng202ilmt}, it does not actually perform
as a standalone language model.
As shown in Equation \ref{eqn:1}, given the label history $\textbf{y}_1^u$, 
the predictor needs to coordinate with the encoder outputs $\textbf{f}_t$ 
to predict the output token $\hat{y}_{t+1}$.
In addition, it also needs to avoid generating repeated label tokens as the duration of each
label normally consists of multiple acoustic frames \cite{ghodsi2020statelessrnn}.
Therefore, the predictor plays a special and important 
role in neural Transducer rather than merely predicting the next vocabulary token as language model.

\begin{figure}[hbtp]
\centering
\includegraphics[width=6.5cm]{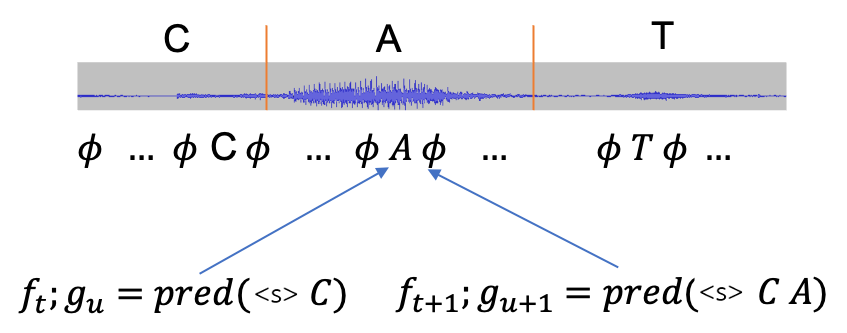}
\caption{An example alignment in neural Transducer for an audio with label ``C A T". The label at time $t$ is $A$ and at time $t+1$ is $\phi$.}
\label{fig:illustration}
\end{figure}

Here, we use a simple example to further illustrate why the predictor is not working 
as a language model. In Figure \ref{fig:illustration}, the label of the
acoustic feature  $\textbf{x}_1^T$ consists of three characters, ``$C\ A\ T$". 
Normally, the acoustic feature sequence is much longer than the label sequence, i.e. $T >> U$.
Figure \ref{fig:illustration} gives an example alignment $\alpha$. 
In the training stage, at the $t$-th frame, the target is ``$A$",
given the encoder output $\textbf{f}_t$ and the label history ``$<$s$>$ $C$".
While at the $t+1$-th frame, the target is $\phi$, given
the encoder output $\textbf{f}_{t+1}$ and label history ``$<$s$>$ $C\ A$".
It is safe to assume that the encoder outputs $\textbf{f}_{t+1}$
and $\textbf{f}_{t}$ are similar since there is only one acoustic
frame difference and they both lie in the middle of the 
pronunciation 
of ``$A$" as illustrated in Figure \ref{fig:illustration}.
In the $t$-th frame, the predictor predicts ``$A$" given the label history 
``$<$s$>$ $C$", which is consistent with the language model task. 
However, in the $t+1$-th frame, the predictor helps to predict ``$\phi$" given label history ``$<$s$>$ $C\ A$", which is different to the language model task where the vocabulary token $T$ is predicted as target. 
Furthermore, there is no blank $\phi$ in the language model.
By considering the large amounts of blank $\phi$ in each alignment,
the predictor is not only working as
a language model. 
More importantly, it needs to coordinate with the acoustic encoder output and label history to generate the neural Transducer alignment.
Therefore, the job 
of predictor is not only predicting normal vocabulary tokens but
also generating the blank $\phi$ for the co-ordination job. 
Because of this reason, the predictor cannot be considered as a pure LM \cite{ghodsi2020statelessrnn}.

\section{Factorized Neural Transducer}
\label{sec:factorized-t-t}

As discussed in Section \ref{subsec:blank}, the predictor has two jobs,
which are predicting the blank and normal vocabulary tokens according to
the Transducer alignment, which hinders the direct use of  text-only data for LM adaptation on 
the predictor. In this section, 
we introduce a novel model architecture, which is called factorized neural Transducer.
Instead of using one predictor to predict both blank and vocabulary tokens, the 
proposed model factorizes the original prediction network into two separate networks, 
predicting vocabulary tokens and blank respectively. A standard
language model could be used
as the vocabulary predictor
since there is no
blank in the output layer. As a result, the vocabulary predictor could be optimized with the 
standard LM loss, i.e. cross entropy, on the label sequence during training, 
and further improved via various LM adaptation techniques given the
target-domain text in test time.

\begin{figure}[hbtp]
\centering
\includegraphics[width=8.5cm]{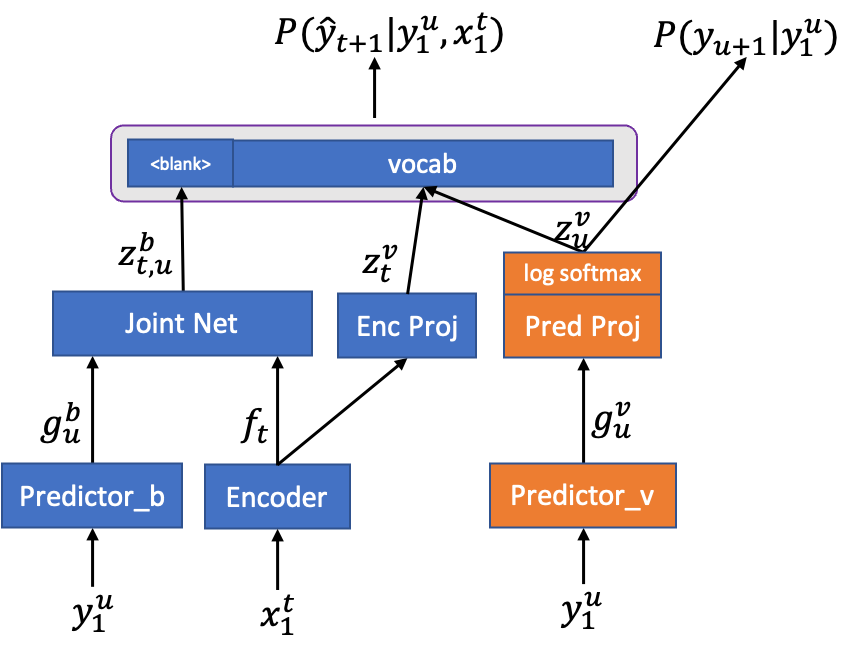}
\caption{Architecture of factorized neural Transducer}
\label{fig:ftt}
\end{figure}

The architecture of the proposed factorized neural Transducer model 
is illustrated in Figure \ref{fig:ftt}.
Two predictors are adopted, one is dedicated to predict the blank $\phi$, which is called blank predictor;
and the other is to predict the label vocabulary excluding $\phi$, which is called vocabulary predictor.
The vocabulary predictor is the same as a normal language model, using history words as input and 
the log probability of each word as output.
The acoustic encoder output $\textbf{f}_t$ is shared by these two predictors 
to extract the acoustic representation, but with slightly
different combinations. For the prediction of the blank $\phi$ , it is important to 
fuse the acoustic and label information as early as possible. Therefore, we adopt the same
combination as \cite{he2019streaming} with a joint network. While for the vocabulary part, we would like to keep
a separate language model module. Hence, the acoustic and label information are 
combined in the logit level, which is similar to the original Transducer 
paper \cite{graves2012sequence}\footnote{we also tried to apply the log\_softmax function on the vocabulary encoder (i.e. Enc Proj) in Figure \ref{fig:ftt}, it presented similar performance, while introduced additional computational cost.}. The exact computation formulas could be
written as below,

\begin{eqnarray}
    \textbf{f}_{t} &=& \text{encoder}(\textbf{x}_1^{t})  \nonumber \\  
    \textbf{g}^{b}_{u} &=& \text{predictor}^{b} (\textbf{y}_1^{u}) \nonumber \\
    \textbf{z}^{b}_{t, u} &=& \textbf{W}_o^b * \text{relu}(\textbf{f}_t+\textbf{g}^{b}_{u})  \nonumber \\
    \textbf{g}^{v}_{u} &=& \text{predictor}^{v} (\textbf{y}_1^{u}) \nonumber \\
    \textbf{z}_t^v &=& \textbf{W}_{enc}^v * \text{relu}(\textbf{f}_t) \nonumber \\
    \textbf{z}_u^v &=& \text{log\_softmax}(\textbf{W}_{pred}^v* \text{relu}(\textbf{g}^{v}_{u})) \nonumber \\
    \textbf{z}^{v}_{t, u} &=& \textbf{z}_t^v + \textbf{z}_u^v \nonumber \\
    P(\hat{y}_{t+1}|\textbf{x}_1^t, \textbf{y}_1^{u}) &=& \text{softmax}([\textbf{z}^b_{t, u};  \textbf{z}^{v}_{t,u}] )
\label{eqn:3}
\end{eqnarray}

The loss function of the factorized Transducer can be written as
\begin{equation}
\mathcal{J}_f = \mathcal{J}_t - \lambda \log P(\textbf{y}_1^U)
\label{eqn:4}
\end{equation}
where the first term is the Transducer loss as defined in Equation \ref{eqn:2}
and the second term is the language model loss with cross entropy. 
$\lambda$ is a hyper-parameter to tune the effect of language model loss. 

The orange part, denoted as vocabulary predictor, in Figure \ref{fig:ftt} has the same structure as a
standalone language model, which can
be viewed as the internal language model in the factorized Transducer. Note that
its output is the log probability over the vocabulary.
Hence in theory this internal language model could be replaced
by any language model trained with the 
same vocabulary, e.g. LSTM and n-gram LMs. 
There is no large matrix computation in the factorized Transducer model in the
joint network as the standard Transducer model. As a result, the training speed and 
memory consumption can be improved compared to the standard
Transducer model, although there is a slight increase of the model 
parameters due to the additional vocabulary predictor.

In the training stage, the factorized Transducer model is trained from scratch 
using the loss function defined in Equation \ref{eqn:4}.
In the adaptation stage, since the vocabulary predictor works as a language
model, we could apply any well-studied language model adaptation techniques to adapt 
the language model on the target-domain text data. For 
simplicity, in this paper, the language model is directly fine-tuned with
the adaptation text data for a specified number of sweeps.

\section{Experiments}
\label{sec:exps}
In this section, we investigate the effectiveness of the 
proposed factorized Transducer model on three test sets, the first one 
is a general test set to verify the impact of architecture change,
and two domain-specific test sets, one is from the public Librispeech corpus, the other is 
an in-house call center task, are used to evaluate the language model adaptation 
of the factorized Transducer.

\subsection{Experiment Setup}
64 thousand (K) hours of transcribed Microsoft data are used as the training data. 
The general test set is used to evaluate the benchmark performance of the
standard and factorized neural Transducer, which covers 13 different application scenarios including dictation, far-field speech and call center, consisting of a total of 1.8 million (M) words.
The word error rate (WER) averaged over all test scenarios is reported.
Two test sets are used for adaptation, the first one is from the public Librispeech data, where
the acoustic transcription of the 960-hour training data is
used as the adaptation text data, and the standard dev and test sets are adopted for evaluation.
The other test set is an internal Microsoft call center test set. 
The statistics of these two test set is summarized in Table \ref{tab:testset}.
All the in-house training and test data are anonymized data with personal
identifiable information removed.
4000 sentence pieces 
trained on the training data transcription was used as vocabulary.
We applied a context window of 8 for the input frames 
to form a 640-dim feature as the input of Transducer encoder
and the frame shift is set to 30ms.
Transformer-Transducer (T-T) models are adopted for all Transducer models,
where the encoder consists of 18 transformer layers and predictor consists of
two LSTM layers. The total number of parameters
for standard and factorized Transducer models are 84M and 103M respectively. Note that the encoder of these two Transducer models are the same and the increase of
model parameter in the factorized Transducer is mainly from the additional 
vocabulary predictor. 
Utterances longer than 30 second were discarded from the 
training data. The T-T models are trained using the chunk based masking
approach as described in \cite{chen2021tt}, with an average latency of
360ms. In this paper, we trained the whole model from scratch without
pretraining on the encoder or predictor.

\begin{table}[t]
    \centering
    \begin{tabular}{c|c|c}
    \hline
       test set     &  adapt words(utts) & test words(utts) \\
    \hline
        Librispeech &    9.4M(281k)        &  210k(11k)     \\ 
        call-center  &   1.4M(76k)        &  77k (4k)     \\
    \hline    
    \end{tabular}
    \caption{Statistics of two test sets for language model adaptation}
    \label{tab:testset}
\end{table}

The first experiment is to evaluate the benchmark performance of the factorized 
T-T model and standard T-T model on the general test set. 
The factorized Transducer with different $\lambda$ as defined 
in Equation \ref{eqn:4} are also investigated.  According to
the results shown in Table \ref{tab:2}, the factorized Transducer
models degrade the WER performance slightly compared to the standard Transducer
model, which is expected due to the factorization of acoustic and label 
information for vocabulary prediction. 
The WER of the factorized T-T increases marginally
with increase of $\lambda$. In contrast, the PPL drops dramatically
when we include the LM loss in Equation \ref{eqn:4} by setting
$\lambda$ larger than 0. An LSTM-LM with same model structure trained 
on the text data results in a PPL of 27.5, which indicates that the 
internal LM extracted from the factorized T-T presents similar PPL compared 
to the standard LM. Note that the PPL reported in Table \ref{tab:2} 
is computed on the sentence piece vocabulary of 4000 tokens.
In the following experiment, we adopt the factorized T-T with $\lambda=0.5$
as the seed model for LM adaptation on text data.

\begin{table}[t]
    \centering
    \begin{tabular}{c|c|c|c}
    \hline
       model        & $\lambda$    & PPL & WER \\
     \hline
      std T-T       &       -        &     &   8.10 \\
    \hline
      \multirow{5}{*}{Factorized T-T}
      &       0.0      &  109.2    &  8.21  \\ 
      &       0.1      &  31.0     & 8.23    \\   
      &       0.2      &  29.3     & 8.25    \\ 
      &       0.5      &  28.0   & 8.32   \\  
      &       1.0      &  27.7   &  8.40  \\
     \hline
    \end{tabular}
    \caption{PPL and WER results of standard and factorized T-T models on 
    the general test set,
    note that PPLs are computed over sentence piece level.}
    \vspace{-0.0cm}
    \label{tab:2}
\end{table}

The next experiment investigates the language model adaptation
of the factorized T-T model on the Librispeech data. 
The experiment results are reported in Table \ref{tab:3}.
Similar to the result on the general test set, the performance of the factorized 
Transducer is slightly worse than the standard Transducer without adaptation.
By using the target-domain text data, significant 
WER reductions can be achieved, resulting
in a relative 15.4\% to 19.4\% WER improvement on the dev and test sets.
The gain over the standard Transducer model is also larger than 10\%
in spite of the improvement of the baseline model. Compared to the WER 
of the shallow fusion on the standard T-T model, which is shown in the second
row of Table \ref{tab:3}, the adapted factorized T-T model still outperforms on most test
sets and gives similar performance on the test clean set.

\begin{table}[t]
    \centering
    \begin{tabular}{c||c||c|c|c|c}
    \hline
    \multirow{3}{*}{model}               &  \multirow{3}{*}{PPL}   & \multicolumn{4}{|c}{WER} \\
    \cline{3-6}
    &                           & \multicolumn{2}{|c}{dev}  & \multicolumn{2}{|c}{test}  \\
    \cline{3-6}
    &                           &clean & other & clean &  other \\
    \hline\hline
    std T-T             &   \multirow{2}{*}{-}     &  5.90 & 13.31 & 5.86 & 13.38 \\
    \cline{0-0} \cline{3-6}
    +shallow fusion    &            &  5.29      &  12.03  &  5.20    &   12.36  \\
    \hline\hline
    factorized T-T      &   88.9     &  6.18 & 13.76 & 6.24 & 14.14    \\ 

    \hline 
    +adapt text             &   \multirow{2}{*}{45.8}     &  4.98 & 11.49 & 5.27 & 11.96   \\  
    (rel. WERR)             &           &  -19.4\% & -16.5\%  & -15.5\% &  -15.4\%     \\
    \hline
    \end{tabular}
    \caption{PPL and WER results of factorized T-T on Librispeech test set for language model adaptation.}
    \label{tab:3}
 \vspace{-0.0cm}
\end{table}

\begin{figure}
    \centering
    \includegraphics[width=8.5cm]{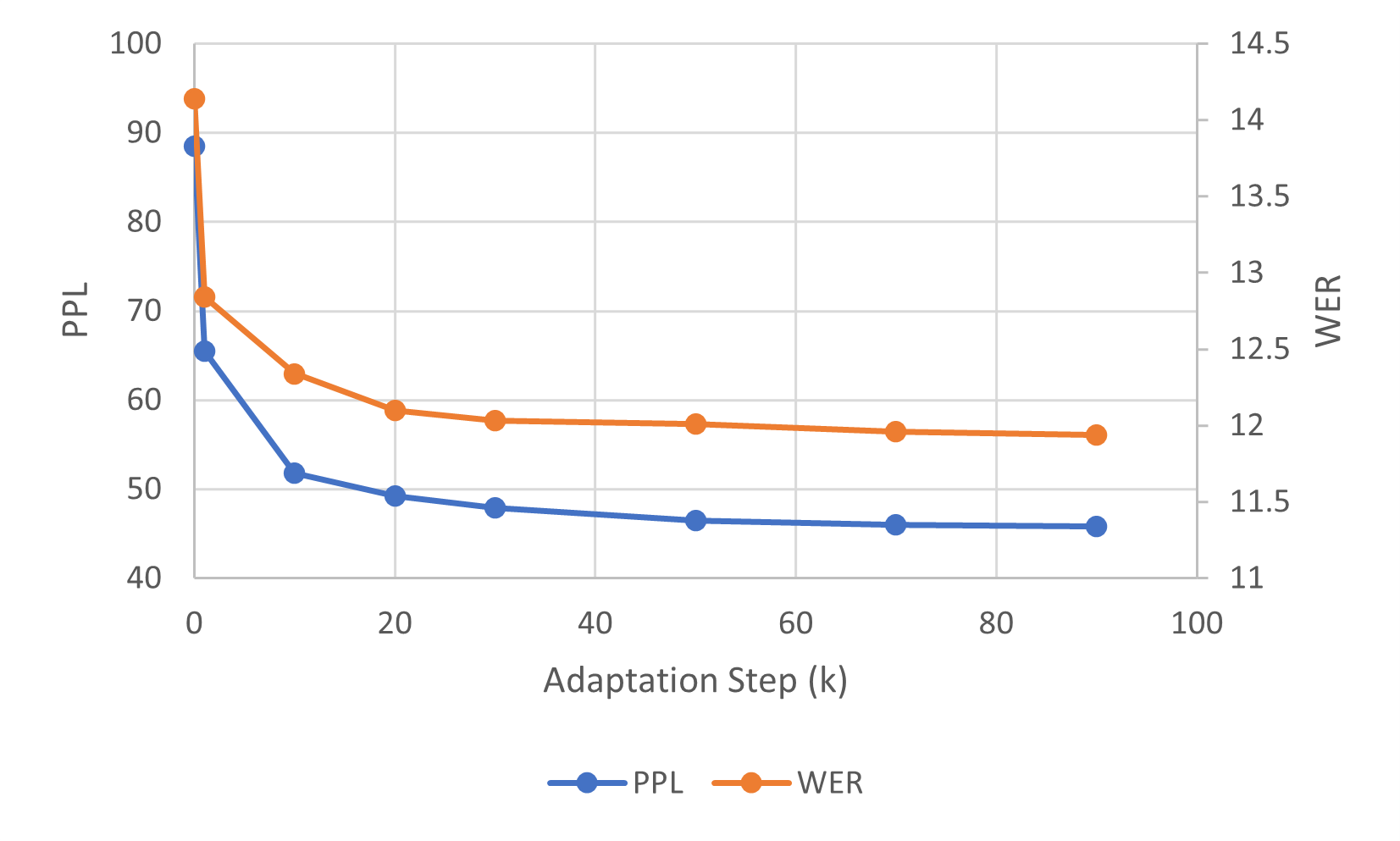}
    \caption{PPL and WER with the increase of LM adaptation step for factorized Transducer in Librispeech test-other set. One sweep of the adaptation text data consists of about 10k steps.}
    \label{fig:5}
\end{figure}

Another experiment is conducted to
reveal the relationship between the improvement of
vocabulary predictor and the performance of factorized Transducer. 
The plot of the PPL and WER trends with
different amounts of adapt text data is given in Figure \ref{fig:5}.
It could be seen that the WER gain of the Transducer  is highly correlated to the PPL improvement of the vocabulary predictor,
which is consistent with the impact of LM in HMM based ASR systems. 
Note that a sweep of the adaptation text data consists of about 10k steps, it could be seen
that four sweeps are enough for the convergence of PPL and WER improvements in the Librispeech task.

The last experiment aims to validate the effectiveness of the language model for 
factorized Transducer on an in-house call center test set. The experiment
results can be found in Table \ref{tab:4}. By fine-tuning the vocabulary predictor
of the factorized Transducer on the adapt text, it yields a relative 15.6\%
WER improvement, which is consistent to the observation on Librispeech, outperforming
the shallow fusion on the standard T-T model.
This demonstrates that the improvement of the 
vocabulary predictor could be transferred to the Transducer model. This is of great practical value
as it is much easier to collect a large scale of text data compared
to the labelled speech data in practice.

\begin{table}[t]
    \centering
    \begin{tabular}{c||c||c}
    \hline
     model   &  PPL  &  WER\\
     \hline\hline
    std T-T &    \multirow{2}{*}{-}    &   37.03  \\
    \cline{0-0} \cline{3-3}
    +shallow fusion &  & 34.36 \\
    \hline\hline
    factorized T-T & 42.0 & 38.14 \\
    \hline
    +adapt         & \multirow{2}{*}{24.1} & 32.18 \\
    (rel. WERR)    &     &  -15.6\%       \\
    \hline
    \end{tabular}
    \caption{PPL and WER results of factorized T-T on the in-house call center test set  for language model adaptation}
    \label{tab:4}
     \vspace{-0.0cm}
\end{table}

\section{Conclusion}
\label{sec:conclusion}
Recent years have witnessed the great success of the E2E based models in speech recognition, especially the
neural Transducer based model due to their streaming capability and promising performance. 
However, how to utilize the out-of-domain text for efficient language model adaptation
remains an active research topic for neural Transducer.
In this paper, we proposed a novel model architecture, factorized neural Transducer,
by separating the blank and vocabulary prediction using two predictor networks,
and adopting a standalone language model as the vocabulary predictor.
Thanks to the factorization, we could adapt the vocabulary predictor with text-only data in the same way as conventional neural language model adaptation. 
The improvement of the language model is able to be transferred to the 
Transducer performance on speech recognition.
The experiment results demonstrate that significant WER improvements
can be achieved by using the target-domain text data, outperforming the shallow fusion on 
standard neural Transducer model.



\vfill\pagebreak
\bibliographystyle{IEEEbib}
\bibliography{refs}

\end{document}